\pdfoutput=1

\documentclass[11pt]{article}

\usepackage[]{acl}

\usepackage{times}
\usepackage{latexsym}
\usepackage{graphicx}
\usepackage{subfig}
\usepackage{booktabs}
\usepackage[T1]{fontenc}

\usepackage[utf8]{inputenc}

\usepackage{microtype}

\usepackage{inconsolata}
\usepackage{amsmath}
\usepackage{tabularx}
\usepackage{graphicx}
\usepackage{xcolor}
\usepackage{multirow}
\usepackage{graphicx}
\usepackage{balance}

\title{Hallucination Diversity-Aware Active Learning for Text Summarization}

\author{Yu Xia$^{1,2}$ \quad Xu Liu$^{1}$ \quad Tong Yu$^{3}$\thanks{\; Corresponding author} \quad Sungchul Kim$^{3}$ \quad Ryan A. Rossi$^{3}$ \\ {\bf Anup Rao$^{3}$ \qquad Tung Mai$^{3}$ \qquad Shuai Li$^{1}$} \\ 
$^{1}$Shanghai Jiao Tong University \qquad $^{2}$University of Michigan \qquad $^{3}$Adobe Research \\
\texttt{xiayuu@umich.edu} \qquad \texttt{\{liu\_skywalker, shuaili8\}@sjtu.edu.cn} \\
\texttt{\{tyu, sukim, ryrossi, anuprao, tumai\}@adobe.com}}

\begin{document}
\maketitle

\begin{abstract}
Large Language Models (LLMs) have shown propensity to generate hallucinated outputs, i.e., texts that are factually incorrect or unsupported. 
Existing methods for alleviating hallucinations typically require costly human annotations to identify and correct hallucinations in LLM outputs. 
Moreover, most of these methods focus on a specific type of hallucination, e.g., entity or token errors, which limits their effectiveness in addressing various types of hallucinations exhibited in LLM outputs.
To our best knowledge, in this paper we propose the first active learning framework to alleviate LLM hallucinations, reducing costly human annotations of hallucination needed. 
By measuring fine-grained hallucinations from errors in semantic frame, discourse and content verifiability in text summarization,  
we propose \textbf{HA}llucination \textbf{D}iversity-\textbf{A}ware \textbf{S}ampling (HADAS) to select diverse hallucinations for annotations in active learning for LLM finetuning.
Extensive experiments on three datasets and different backbone models demonstrate advantages of our method in effectively and efficiently mitigating LLM hallucinations.
\end{abstract}
\section{Introduction}\label{sec:intro}

Despite the prominent capabilities of large language models (LLMs) in natural language generation (NLG) tasks~\cite{brown2020language, chung2022scaling, touvron2023llama}, a notable limitation of them lies in their propensity to hallucinate~\cite{ji2023survey, manakul-etal-2023-selfcheckgpt, zhao2023survey, bang-etal-2023-multitask, peng2023check}, where models generate seemingly plausible but ungrounded outputs that either contradict or cannot be verified by existing sources. 
The phenomenon of hallucination poses a crucial challenge to the real-world applications of LLMs, where the models' faithfulness and trustworthiness are emphasized~\cite{yang2023harnessing, zhao2023survey}.

While many methods have been proposed recently to detect hallucinations in the outputs of LLMs~\cite{manakul-etal-2023-selfcheckgpt, li-etal-2023-halueval, mundler2023self}, how to efficiently and effectively alleviate hallucinations in LLMs remains a notable problem.
Existing methods for hallucination mitigation often focus on finetuning LLMs with human feedback or human-annotated samples to align the models' outputs with human-plausible content~\cite{ouyang2022training, sun2022contrastive, wu2023fine}.
While these methods have proven effective, they often require large amounts of costly human annotations to identify and rectify hallucinations in LLM outputs~\cite{zhao2023survey, guerreiro2023hallucinations, perlitz2023active, xia2024llm}. Moreover, most of them emphasize mitigating a specific type of hallucination, e.g., entity or token errors \cite{nan-etal-2021-entity, cao-etal-2022-hallucinated}, which limits their applicability in addressing various types of hallucinations comprehensively.


Aiming to reduce the intensive amount of human annotations needed, in this paper we propose an active learning framework to finetune LLMs for hallucination mitigation. In this framework, we actively select samples that LLMs may hallucinate on for annotation and finetuning.
As the text summarization task has gained wide attention in factuality evaluations, which measure whether the model's outputs are faithful to the source document~\cite{ maynez2020faithfulness, pagnoni2021understanding, ji2023survey, li-etal-2023-halueval, zhang2023benchmarking}, we instantiate our active learning framework to address LLM hallucinations in generated summaries. 
We revisit the different types of hallucinations in text summarization defined by \citet{pagnoni2021understanding}. Then, we leverage corresponding detection models \cite{zhang2020bertscore, zhong2022towards, feng-etal-2023-factkb} to measure fine-grained hallucinations, including semantic frame errors, discourse errors, and content verifiability errors, for annotation sample selection.

While measuring potential hallucinations of all three types, greedily choosing the samples most likely to exhibit hallucinations may result in an excessive focus on addressing a certain type of hallucination while overlooking others. 
For example, if the evaluation score for semantic frame hallucinations dominates among all three types, greedy selection would then lead to choosing samples that mostly result in semantic frame errors for human annotations. As a result, the finetuned LLMs may reduce semantic frame hallucinations effectively but still suffer from other types. 
To address this limitation and take into account the diversity of hallucination samples, we propose a sample selection strategy for our active learning framework, called \textbf{HA}llucination \textbf{D}iversity-\textbf{A}ware \textbf{S}ampling (HADAS).
Extensive experiments demonstrate the advantage of our proposed method in alleviating hallucinations, while also limiting the amount of costly human annotations, compared with both the random sampling baseline and the existing sample selection approaches for text summarization.

In summary, we make the following contributions in this work: i) To our best knowledge, in this paper we propose the first active learning framework to alleviate LLM hallucinations, reducing the amount of human annotations needed; ii) We propose a sample selection strategy HADAS to select samples of diverse hallucination types; iii) We demonstrate with extensive experiments the effectiveness of our proposed active learning method in mitigating hallucinations in text summarization.


\section{Related Work}
\subsection{Hallucination Mitigation in LLM}

The hallucination problem has been a pressing topic in recent studies on LLMs, where models generate incorrect or non-existent information that either contradicts or is unsupported by existing sources~\cite{ji2023survey, yang2023harnessing}. 
Although there is a growing number of studies on hallucination detection and evaluation in LLMs \cite{manakul-etal-2023-selfcheckgpt, bang-etal-2023-multitask, guerreiro2023hallucinations, li-etal-2023-halueval, mundler2023self}, how to effectively and efficiently mitigate hallucinations remains a notable challenge.
A few recent works have explored addressing the hallucination problem during inference time via improved decoding strategies \cite{lee2022factuality, shi2023trusting, wan-etal-2023-faithfulness}, retrieval augmentation \cite{shuster-etal-2021-retrieval-augmentation, peng2023check}, and self-verification \cite{varshney2023stitch, dhuliawala2023chain}.
Another line of works focus on finetuning LLMs to hallucinate less with various learning paradigms.
\citet{wan-bansal-2022-factpegasus} incorporate factual consistency as one of the training objectives during finetuning.
\citet{sun2022contrastive} use contrastive learning to reduce hallucination by comparing faithful samples with hallucinated samples.  
\citet{roit-etal-2023-factually} leverage reinforcement learning to align LLMs' outputs to be more factually consistent to the source document with a natural language inference model. 
While these methods have been proven effective, they typically require a large amount of costly human annotations. 
In comparison, our proposed active learning framework for LLM finetuning aims to mitigate hallucinations while minimizing the amount of human annotations needed.



\begin{table*}[!t]
\fontsize{9pt}{11pt}\selectfont

\renewcommand{\arraystretch}{1.25} 
    \begin{tabularx}{\textwidth}{XX}
    \toprule
    \textbf{Source Document} & \\
    \hline
    \multicolumn{2}{p{\dimexpr\textwidth-0.4cm}}{Heavy rains and flooding have forced hundreds of thousands of people from homes in southern Mexico's state of Tabasco over the past four days, with nearly as many trapped by the rising waters, state officials said Thursday. Officials say about 300,000 people are still trapped by the worst flooding in the region for 50 years ...}
    \\ \toprule
    \textbf{Hallucination Type} & \textbf{Example Summary} \\ \hline
    \textit{Semantic Frame Error:}
    The entity or predicate in the summary is not inconsistent with source document. & Recent heavy rains in \textcolor{red}{northern} Mexico have caused the worst flooding in 50 years.
    \\ \hline
    \textit{Discourse Error:}
    The statements or references in the summary are linked in an erroneous way. & \textcolor{red}{Due to} the worst flooding in 50 years in Tabasco, officials report that heavy rains began last Thursday.  \\ \hline
    \textit{Content Verifiability Error:}
    The information in the summary is not present or verifiable in source document.  & Due to heavy rains in southern Mexico, \textcolor{red}{a state emergency was declared} in Tabasco.
    \\ \bottomrule
    \end{tabularx}
    \caption{Examples of three types of hallucinations in text summarization following the typology proposed by \citet{pagnoni2021understanding}. The source document is from CNN-DailyMail \cite{cnndm2015}. We highlight the hallucinated content in red.}
    \label{tab:hallucination example}
    \vspace{-0.5em}
\renewcommand{\arraystretch}{1.0}
\end{table*}

\subsection{Active Learning in NLG}


Active learning is a well-known technique employed in natural language processing to reduce annotation efforts by actively selecting informative samples~\cite{zhang-etal-2022-survey}. In the context of language modeling, active learning is mainly used for text classification tasks \cite{ein-dor-etal-2020-active, margatina-etal-2021-active, wu-etal-2022-context, yu-etal-2022-actune}, such as named entity recognition \cite{shen2018deep, radmard2021subsequence}. A few recent works have explored active learning methods for NLG tasks.
\citet{tsvigun2022active} propose the first effective diversity-based active learning query strategy for text summarization based on the embedding similarities between source documents. The authors report that the uncertainty-based strategy does not perform well and is outperformed by the random sampling baseline.
\citet{perlitz2023active} evaluate the performance of existing active learning strategies across various NLG tasks such as paraphrase generation, summarization, and question generation. The authors suggest that compared to classification tasks, the lack of clearly defined ground-truth labels in NLG tasks poses difficulties in measuring uncertainty, which contributes to poor performance in uncertainty-based sample selection.
As LLMs' hallucinations typically occur in NLG tasks, applying active learning for hallucination mitigation is an unexplored and non-trivial task. Our work proposes a diversity-based sampling strategy addressing LLMs' hallucinations in text summarization. 
Note that while \citet{tsvigun2022active} also proposes a diversity-based method for text summarization, it aims to select document samples that are semantically diverse. 
In contrast, the diversity considered in our method focuses on various types of hallucinations in generated summaries. 
Thus, we make the first attempt towards an active learning paradigm for hallucination mitigation in NLG.


\section{Hallucination Typology Revisit}\label{sec:halu_examples}


Since LLMs may hallucinate in different forms \cite{ji2023survey}, evaluations of hallucination have received increasing attention in recent studies. Particularly, a variety of factuality metrics have been developed~\cite{maynez2020faithfulness, wang2020asking, pagnoni2021understanding, zhong2022towards, fabbri2022qafacteval, feng-etal-2023-factkb}, including entity or token hallucination \cite{liu-etal-2022-token}, sentence hallucination \cite{manakul-etal-2023-selfcheckgpt}, and relation hallucination \cite{zha-etal-2023-alignscore}.

Aiming to comprehensively mitigate hallucinations in text summarization, we follow the typology proposed in \citet{pagnoni2021understanding} and introduce three types of common hallucinations in text summarization: i) \textit{Semantic Frame Error}, ii) \textit{Discourse Error}, and iii) \textit{Content Verifiability Error}.
To provide more background information, we provide illustrative examples of these three types of hallucination in Table~\ref{tab:hallucination example}.
Specifically, given a news article reporting the heavy rains and flooding in the southern Mexico area as the source document, a semantic frame error in the summary can be an entity or predicate incorrectly interpreted from the source, e.g. southern being misinterpreted as northern as illustrated in the first example summary.
Additionally, a discourse error refers to the case when statements or claims in a sentence are linked together erroneously in terms of temporal ordering or causal links, e.g. the second example summary mistakenly states that the flooding was the cause of heavy rains.
Moreover, a content verifiability error stands for the extrinsic information unverifiable from the source document, e.g., the declared state emergency in the third example summary is not mentioned in the source text.
These examples demonstrate the varying forms of hallucinations that LLMs may generate in the summary.

Hence, by evaluating hallucinations in these various aspects, we can derive a more comprehensive understanding of hallucinations in LLM text summarization.
This motivates us to capture and mitigate diverse types of hallucinations and enhance the faithfulness of LLMs in text summarization.

\begin{figure*}[!t]
    \centering
    \includegraphics[width=0.98\linewidth]{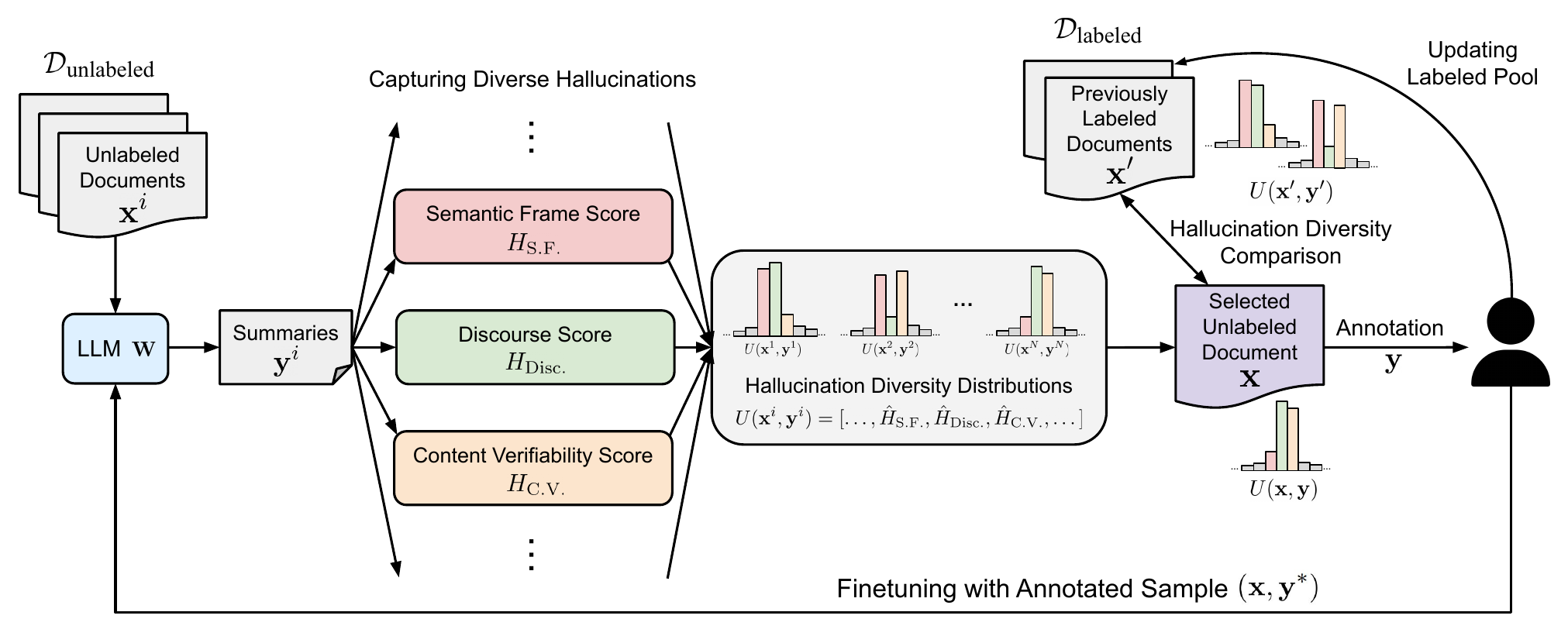}
    \vspace{-0.5em}
    \caption{Overview of our hallucination diversity-aware active learning framework.
    }
    \vspace{-0.5em}
    \label{fig:framework}
\end{figure*}

\section{Methodology}

We first present our proposed active learning framework for LLM finetuning in hallucination mitigation in Section \ref{sec:al_tune}. Section \ref{sec:halu_measurements} details how we capture diverse hallucination types using off-the-shelf detection models. Then in Section \ref{sec:hadas}, we describe in detail our hallucination diversity-aware sample selection strategy. 

\subsection{Active Learning for LLM Finetuning}\label{sec:al_tune}

We formulate our proposed active learning framework for LLM finetuning in text summarization with a feedback loop between the LLM and the annotator, as illustrated in Figure 1. 
We first introduce the necessary notations as follows.

Given an LLM, we denote its weights as $\mathbf{W}$. 
We denote an input document as $\mathbf{x}=\left(x_1 \ldots x_m\right)$ and the summary generated by the LLM as $\mathbf{y}=\left(y_1 \ldots y_n\right)$, where $m$ and $n$ are the token lengths of the document and generated summary respectively. 
Suppose we have a total of $N$ documents in the unlabeled pool, denoted as $\mathcal{D}_{\text{unlabeled}} = \{\mathbf{x}^i\}_{i=0}^N$. 
An unlabeled document means that no annotation is currently available to identify and correct the potential hallucination in LLM-generated summary $\mathbf{y}$ based on this document. 
We also keep track of a labeled pool $\mathcal{D}_{\text{labeled}} = \{(\mathbf{x}^j, {\mathbf{y}^*}^j)\}_{j=0}^M$, where $\mathbf{y}^*$ denotes the annotated summary and $M$ is the size of labeled pool. We denote a sample selection strategy as $\mathcal{A}$ and the active learning loop consists of the following three main steps.

\paragraph{Sample Selection.}
To select a document $\mathbf{x}^i$ from the unlabeled pool $\mathcal{D}_{\text{unlabeled}}$, the LLM with weights $\mathbf{W}$ first generates a summary $\mathbf{y}^i$ for each document.
Then, based on the selection strategy instantiated by the query function $\mathcal{A}$, we choose
\begin{equation}\label{eq:A}
(\mathbf{x}, \mathbf{y})=\arg\max _{i \in {1, \ldots, N}} \mathcal{A}((\mathbf{x}^i, \mathbf{y}^i) \mid \mathcal{D_{\text{unlabeled}}}, \mathbf{W})\;, 
\end{equation}
which maximizes the designed criteria of $\mathcal{A}$ to choose the most informative samples for hallucination mitigation.

\paragraph{Human Annotation.} The selected document-summary pair $(\mathbf{x}, \mathbf{y})$ is then annotated by examining and correcting $\mathbf{y}$ for hallucinated content based on the source document $\mathbf{x}$. The annotated summary denoted as $\mathbf{y}^*$ is collected. Subsequently, the document $\mathbf{x}$ is removed from the unlabeled pool and added to the labeled pool along with $\mathbf{y}^*$:
$$\mathcal{D}_{\text{unlabeled}}:=\mathcal{D}_{\text{unlabeled}}\;\backslash\;\{\mathbf{x}\}\;,$$
$$\mathcal{D}_{\text{labeled}}:=\mathcal{D}_{\text{labeled}}\cup\{(\mathbf{x}, \mathbf{y}^*)\}\;.$$

\paragraph{Model Finetuning.} After receiving the annotated document-summary pair $(\mathbf{x}, \mathbf{y}^*)$, the LLM is finetuned, and its weights are updated based on the document and the hallucination-annotated summary:
\begin{equation*}
    \hat{\mathbf{W}}=\arg \min_{\mathbf{W}} \mathcal{L}((\mathbf{x}, \mathbf{y}^*), \mathbf{W})\;,
\end{equation*}
where $\mathcal{L}$ is the loss function used for LLM finetuning, e.g., supervised finetuning objective. Then the updated LLM is evaluated on the validation or test set. 
Next, the LLM with the updated weights 
$\hat{\mathbf{W}}$ is used for the new round of sample selection with similar procedures as described previously.

Such iterative learning process is repeated until a stopping criterion is met, such as reaching a preset number of iterations or when the model's performance on the validation set no longer improves after a certain number of consecutive rounds.

\subsection{Capturing Diverse Hallucination Types}\label{sec:halu_measurements}
As discussed in Section \ref{sec:halu_examples}, hallucinations in LLMs can be of various types. 
To select samples that LLMs tend to hallucinate on, we aim to capture different types of hallucination in text summarization. Specifically, we adopt three hallucination detection methods measuring semantic frame errors, discourse errors, and content verifiability errors, respectively. The details are described as follows.

Note that we are fully aware that there are many emerging new types of hallucinations beyond the three types we considered here. 
As it is unrealistic to exhaustively take into account all hallucination evaluation methods, we follow a well-defined typology proposed by \citet{pagnoni2021understanding} to capture three common hallucinations in text summarization.  
Our contribution lies in developing a generic active learning framework for hallucination mitigation, which offers flexibility in that new measurements of hallucinations can be easily integrated.


\paragraph{Semantic Frame Score $H_{\mathrm{S.F.}}$.}
As suggested and validated in \citet{pagnoni2021understanding}, \citet{ribeiro-etal-2022-factgraph}, and \citet{feng-etal-2023-factkb}, entailment-based models show clear advantages in detecting hallucinations on semantic frames, due to their fine-grained representation of facts, entities, and relations. 
Therefore, we adopt a recent entailment-based model FactKB~\cite{feng-etal-2023-factkb} to evaluate semantic frame errors, which achieves state-of-the-art performances on factual consistency detection and high correlations with human judgments. 
The model takes the document-summary pair as input and outputs a probability of the summary being factually consistent with the document, which we denote as the semantic frame (S.F.) score $H_{\mathrm{S.F.}}$.

\paragraph{Discourse Score $H_{\mathrm{Disc.}}$.}
Different from semantic frames that focus on parts of a sentence like entities, detecting discourse errors such as erroneously connected claims requires a view of the entire sentence \cite{pagnoni2021understanding}.
Therefore, sentence-level detection which is widely adopted in recent QA-based models \cite{wang2020asking, fabbri2022qafacteval, zhong2022towards} comes in handy.
The idea behind these methods is to compose each sentence of the models' outputs as a question and then ask a pre-trained QA model to answer whether this sentence is faithful to the source document.
We adopt a recent QA-based method, UniEval \cite{zhong2022towards}, which leverages a pretrained T5 model, further enhancing its natural language understanding ability at the sentence level.
We denote the probability of the model answering ``Yes'' to the question as the discourse score $H_{\mathrm{Disc.}}$.


\paragraph{Content Verifiability Score $H_{\mathrm{C.V.}}$.} 
For content verifiability, the main goal is to evaluate whether the information in the summary is present in the source document.
Thus, as observed by \citet{pagnoni2021understanding}, \citet{ribeiro-etal-2022-factgraph}, and \citet{feng-etal-2023-factkb}, token-level similarity metrics such as BERTScore \cite{zhang2020bertscore} perform competitively well.
Therefore, we choose BERTScore-Precision (BERT-P), which is more correlated with human judgments according to \citet{feng-etal-2023-factkb}, as our content verifiability score denoted as $H_{\mathrm{C.V.}}$.

\subsection{Hallucination Diversity-Aware Sampling}\label{sec:hadas}
In this section, we describe in detail our proposed sample selection strategy that selects diverse hallucination samples for annotations. 

\paragraph{Hallucination Score $H_{\mathrm{Halu.}}$.}

With the above scores focusing on three different hallucination types, a natural idea for active learning sample selection strategy is to greedily select samples with the lowest total hallucination scores. 
Given a sample of document-summary pair $(\mathbf{x}, \mathbf{y})$, the hallucination score for this sample is calculated as
\begin{equation}\label{eq:halu}
H_{\mathrm{Halu.}}(\mathbf{x}, \mathbf{y}) = w_1\hat{H}_{\mathrm{S.F.}} + w_2\hat{H}_{\mathrm{Disc.}} + w_3\hat{H}_{\mathrm{C.V.}} 
\end{equation}
where $\hat{H}_{\mathrm{S.F.}}$ is the min-max normalized value of $H_{\mathrm{S.F.}}$ and similarly for $\hat{H}_{\mathrm{Disc.}}$ and $\hat{H}_{\mathrm{C.V.}}$, and $w_1$, $w_2$, and $w_3$ are the weights for each score. Note that for the three hallucination scores discussed in Section \ref{sec:halu_measurements}, the higher the score, the better. Thus, the lower the $H_{\mathrm{Halu.}}$, the more hallucinations occur in the generated summary. 

Such greedy exploitation of hallucination scores, however, might not lead to the most informative sample selections, as it might give excessive focus on a certain type of hallucination. For example, if semantic frame errors are more common and scores of $H_{\mathrm{S.F.}}$ are consistently low, the hallucination score $H_{\mathrm{Halu.}}$ would be predominantly influenced by $H_{\mathrm{S.F.}}$. This could result in the selection of samples primarily exhibiting semantic frame errors, while neglecting other types of hallucinations.

\begin{table*}[!t]
\centering
\fontsize{7.21pt}{8.6pt}\selectfont
\tabcolsep=0.105cm
\renewcommand{\arraystretch}{1.3} 
\begin{tabular}{ll|cccc|cccc|cccc}
\hline
& & \multicolumn{4}{c}{\textbf{CNN-DailyMail}} & \multicolumn{4}{c}{\textbf{Multi-News}} & \multicolumn{4}{c}{\textbf{Gigaword}} \\
\cline{3-14}
\textbf{Model} & \textbf{Method} & BERT-P & UniEval & FactKB & ROUGE-L & BERT-P & UniEval & FactKB & ROUGE-L & BERT-P & UniEval & FactKB & ROUGE-L \\
\hline
\hline
\multirow{4}{*}{\begin{tabular}{@{}c@{}}Flan-T5 \\ Small\end{tabular}}
& Random & 73.30 & 60.12 & 69.83 & 13.76 & 67.84 & 46.68 & 62.60 & 9.63 & 56.71 & 36.72 & 7.50 & 23.06 \\
& IDDS & 74.92 & 63.96 & 76.58 & 14.63 & 66.96 & 50.06 & 66.00 & 9.40 & \underline{57.42} & \underline{39.22} & 9.00 & \underline{23.67}  \\
& HADAS$_{\text{w/o\;Div.}}$ & \underline{76.64} & \underline{70.63} & \underline{82.26} & \underline{15.36} & \underline{68.95} & \underline{50.66} & \underline{68.49} & \underline{10.08} & 57.40 & 33.77 & \underline{9.23} & 22.29  \\
& \textbf{HADAS} & \textbf{78.63} & \textbf{75.75} & \textbf{87.46} & \textbf{16.55} & \textbf{70.26} & \textbf{56.40} & \textbf{74.22} & \textbf{11.04} & \textbf{61.06} & \textbf{40.53} & \textbf{10.85} & \textbf{23.89}  \\
\hline 
\hline
\multirow{4}{*}{\begin{tabular}{@{}c@{}}Flan-T5 \\ Base\end{tabular}}
& Random & 69.26 & 58.65 & 69.25 & 15.12 & 65.51 & 47.71 & 52.69 & 7.45 & 56.33 & 42.00 & \underline{7.33} & 27.29 \\
& IDDS & 70.64 & 63.95 & 74.22 & 15.42 & 62.22 & 40.17 & 41.74 & 6.68 & \underline{56.77} & \underline{43.97} & 6.49 & 27.09  \\
& HADAS$_{\textrm{w/o\;Div.}}$ & \underline{72.05} & \underline{67.42} & \underline{77.13} & \underline{16.51} & \underline{69.83} & \underline{56.82} & \underline{61.57} & \underline{9.33} & 54.93 & 39.10 & 5.43 & \underline{28.39}  \\
& \textbf{HADAS} & \textbf{73.74} & \textbf{70.31} & \textbf{80.73} & \textbf{17.19} & \textbf{70.82} & \textbf{61.12} & \textbf{66.39} & \textbf{9.87} & \textbf{59.36} & \textbf{47.98} & \textbf{9.18} & \textbf{29.25}  \\
\hline 
\hline
\multirow{4}{*}{\begin{tabular}{@{}c@{}}BART \\ Base\end{tabular}}
& Random & 76.08 & 74.02 & 89.65 & 19.57 & 69.25 & 50.52 & 76.72 & 12.78 & 79.78 & 61.23 & 51.08 & 35.32  \\
& IDDS & 74.25 & 68.01 & 88.86 & 19.39 & \underline{71.00} & \underline{53.49} & \underline{80.06} & \underline{14.80} & 83.63 & \underline{62.71} & 55.43 & \underline{35.56}  \\
& HADAS$_{\textrm{w/o\;Div.}}$ & \underline{77.56} & \underline{75.42} & \underline{92.82} & \underline{19.95} & 68.68 & 50.78 & 75.81 & 13.28 & \underline{85.59} & 56.60 & \underline{69.44} & 35.11  \\
& \textbf{HADAS} & \textbf{78.14} & \textbf{76.65} & \textbf{93.95} & \textbf{20.12} & \textbf{71.03} & \textbf{55.94} & \textbf{80.22} & \textbf{14.83} & \textbf{87.59} & \textbf{63.75} & \textbf{70.12} & \textbf{35.91}  \\
\hline 
\end{tabular}
\renewcommand{\arraystretch}{1.0}
\caption{Main results of summarization factuality and quality metrics with 30\% of hallucination annotations across models and datasets, where the best results are highlighted in bold, and the second-best are underscored.}
\label{tab:main}
\vspace{-0.5em}
\end{table*}

\paragraph{Hallucination Diversity Score $H_{\mathrm{Div.}}$.}

To further address the limitation of the greedy method, we propose a hallucination diversity-based sample selection strategy, \textbf{HA}llucination \textbf{D}iversity-\textbf{A}ware \textbf{S}ampling (HADAS). 
The main idea behind HADAS is that it would query the samples that have low hallucination scores while ensuring at the same time the hallucination types of selected samples as dissimilar (i.e., diverse) as possible.

To measure the similarity between hallucinations, we consider normalized scores of hallucination types as hallucination distribution $U$ as 
$$U(\mathbf{x}, \mathbf{y}) = [\hat{H}_{\mathrm{S.F.}}, \hat{H}_{\mathrm{Disc.}}, \hat{H}_{\mathrm{C.V.}}]\;,$$
where additional hallucination metrics on other types can be easily included as illustrated in Figure \ref{fig:framework}.
Then, we calculate the average Jensen-Shannon Divergence between the hallucination distribution of each unlabeled sample and all samples in the labeled pool as the diversity score $H_{\mathrm{Div.}}$. 
Formally, given a unlabeled document and LLM-generated summary $(\mathbf{x}, \mathbf{y})$, its diversity score is calculated as
\begin{equation}\label{eq:div}
H_{\mathrm{Div.}}(\mathbf{x}, \mathbf{y}) = \frac{\sum_{\mathcal{D}_{\text{labeled}}} \texttt{JSD}(U(\mathbf{x},\mathbf{y}), U(\mathbf{x^{\prime}}, \mathbf{y^{\prime}}))}{|\mathcal{D}_{\text{labeled}}|}\,,
\end{equation}
where $(\mathbf{x^{\prime}}, \mathbf{y^{\prime}})$ are samples from the labeled pool $\mathcal{D}_{\text{labeled}}$, and \texttt{JSD} represents the Jensen-Shannon Divergence measure. A higher $H_{\mathrm{Div.}}$ value indicates higher diversity of the unlabeled sample $(\mathbf{x}, \mathbf{y})$ compared to previously labeled samples.

With the hallucination score defined in Equation \ref{eq:halu} and the diversity score defined in Equation \ref{eq:div}, we propose the following query function $\mathcal{A}$ that implements selection criteria in Equation \ref{eq:A}:
\begin{equation}
    \mathcal{A}(\mathbf{x}, \mathbf{y}) = \lambda H_{\mathrm{Div.}}(\mathbf{x}, \mathbf{y}) - (1 - \lambda) H_{\mathrm{Halu.}}(\mathbf{x}, \mathbf{y})\,,
\end{equation}
where $\lambda \in [0,1]$ is a hyperparameter.
With the sample selection strategy implemented, we have completed the active learning framework for hallucination mitigation as formulated in Section \ref{sec:al_tune} and illustrated in Figure \ref{fig:framework}.


\section{Experiment Setup}\label{sec:experiment}

\subsection{Backbone Models}\label{sec:models}
We conduct our experiments mainly on three backbone LLMs, Flan-T5 Small~\cite{chung2022scaling}, Flan-T5 Base~\cite{chung2022scaling}, and BART Base~\cite{lewis-etal-2020-bart}. The models are selected following 
\cite{tsvigun2022active} and based on their distinctive strengths in text summarization. For Flan-T5 Small and Flan-T5 Base, we directly prompt the models with the instruction \texttt{``Summarize:''} as they have been instruction-tuned to do summarization task. For BART Base, we follow \citet{lewis-etal-2020-bart} to use a BART Base model finetuned on XSum dataset to ensure the summarization quality. 

\subsection{Datasets and Metrics}
We choose three datasets, CNN-DailyMail \cite{cnndm2015}, Multi-News \cite{alex2019multinews}, and Gigaword \cite{rush-etal-2015-neural}. 
For computational efficiency, following \citet{shen2018deep} and \citet{tsvigun2022active}, we select a subset of samples from each dataset.
Specifically, for CNN-DailyMail dataset, we randomly sample 5,000 samples from the training set, 500 from the test set, and 250 from the validation set. 
For both Multi-News and Gigaword datasets, we first randomly sample 2,000 samples from the training set. 
To better demonstrate improvements on these two datasets, we intentionally apply filtering to choose 200 samples from the test set and 100 samples from the validation set that are more prone to hallucinations by the models, as measured by the metrics introduced in Section \ref{sec:halu_measurements}, with $H_{\mathrm{C.V.}}$ lower than 60 and both $H_{\mathrm{Disc.}}$ and $H_{\mathrm{S.F.}}$ lower than 40.

To evaluate the performance of our methods, we use the three hallucination detection metrics as introduced in Section \ref{sec:halu_measurements}: FactKB~\cite{feng-etal-2023-factkb} for semantic frame, UniEval~\cite{zhong2022towards} for discourse, and BERT-P~\cite{zhang2020bertscore} for content verifiability. In addition, we also measure the ROUGE-L~\cite{lin2004rouge} score to assess the quality of generated summaries.

\begin{figure*}[t!]
    \centering
    \subfloat[FactKB]{\includegraphics[width=0.248\textwidth]{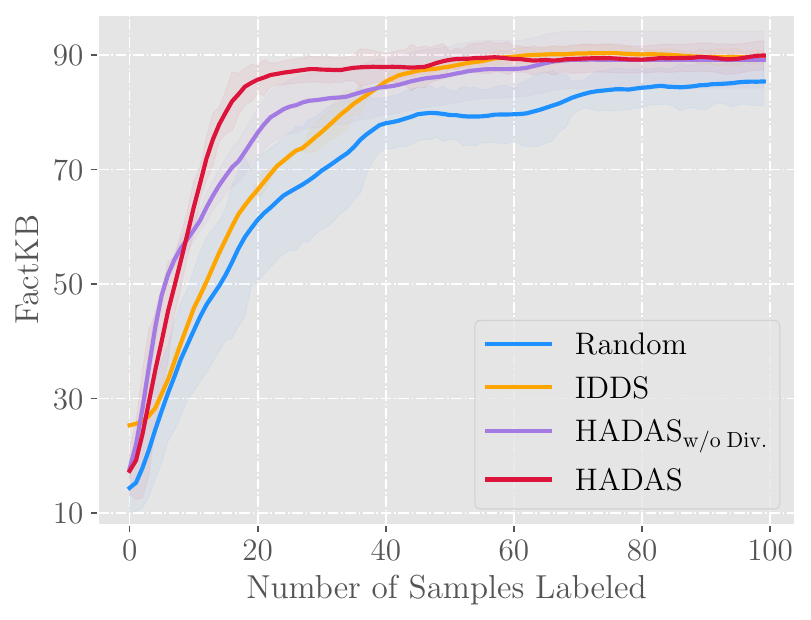}}
    \subfloat[UniEval]{\includegraphics[width=0.248\textwidth]{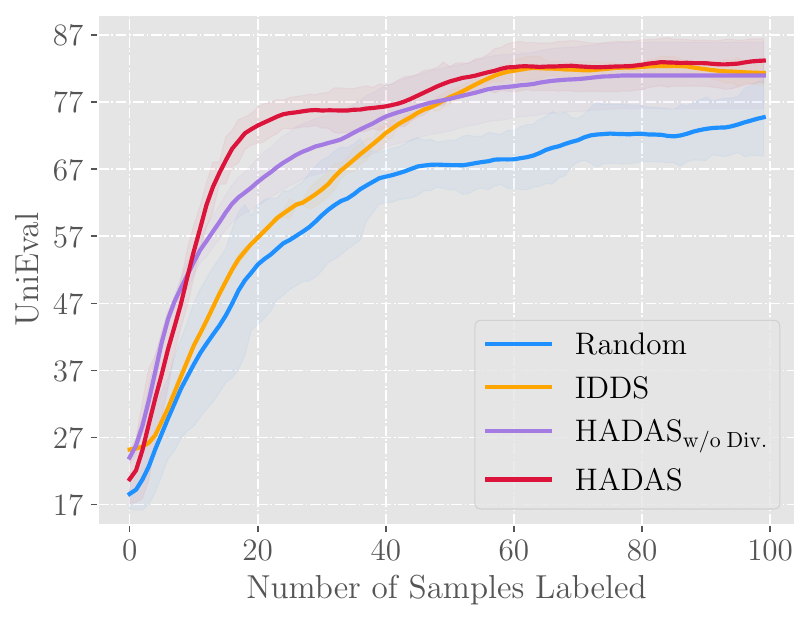}}
    \subfloat[BERT-P]{\includegraphics[width=0.248\textwidth]{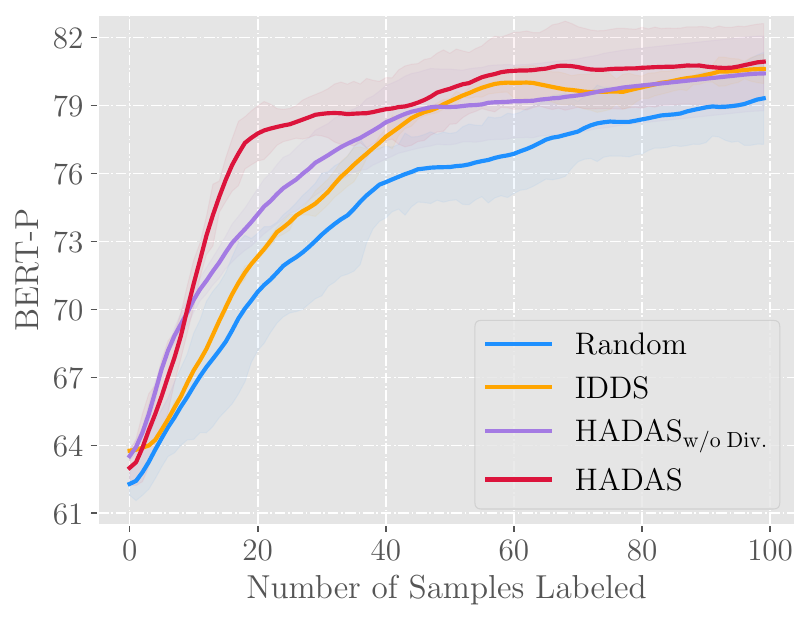}}
    \subfloat[ROUGE-L]{\includegraphics[width=0.248\textwidth]{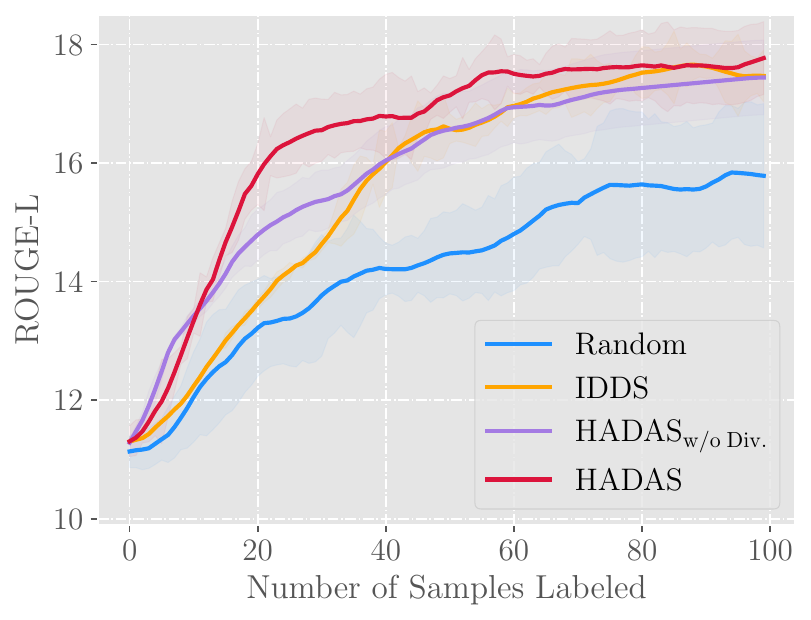}}
    \vspace{-0.5em}
    \caption{Factuality and quality curves over full hallucination annotations of Flan-T5 Small on CNN-DailyMail.}
    \label{fig:cnn-flan-t5-small} 
    \vspace{-1em}
\end{figure*}
\begin{figure*}[t!]
    \centering
    \subfloat[FactKB]{\includegraphics[width=0.248\textwidth]{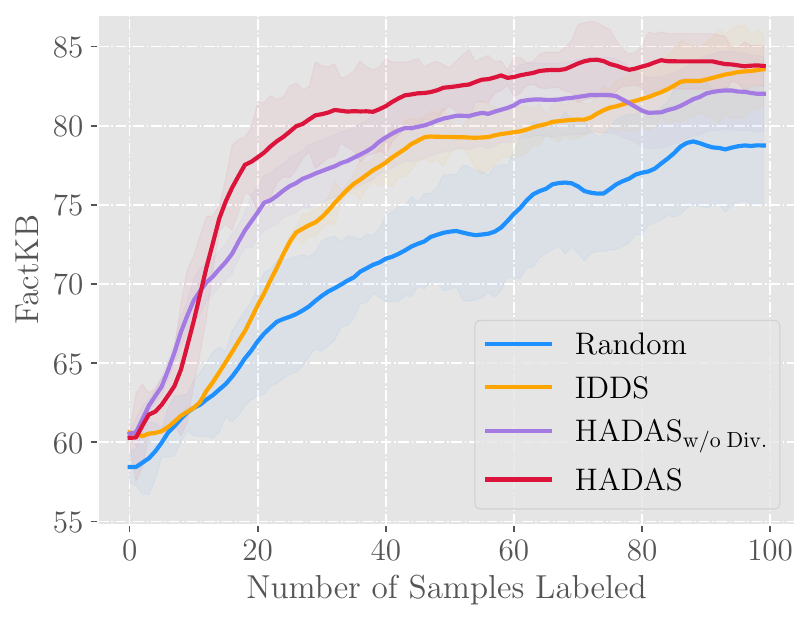}}
    \subfloat[UniEval]{\includegraphics[width=0.248\textwidth]{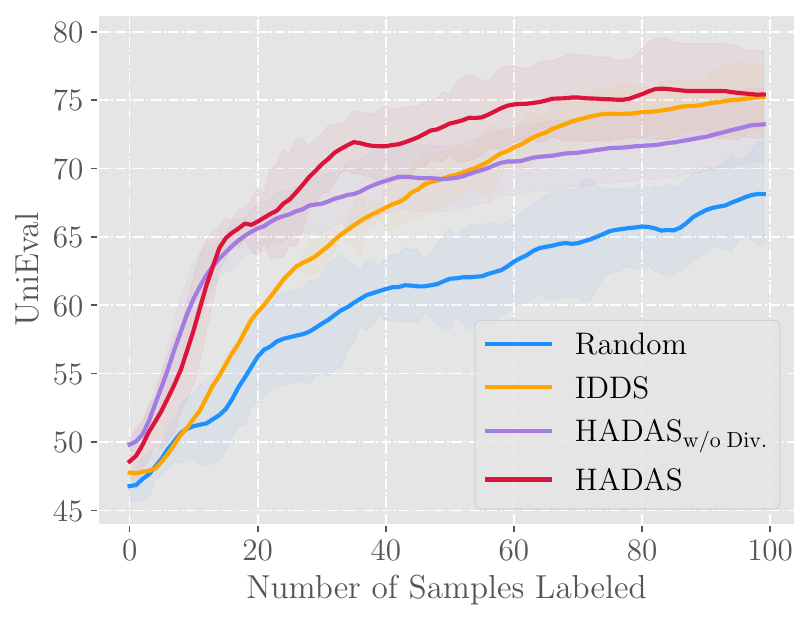}}
    \subfloat[BERT-P]{\includegraphics[width=0.248\textwidth]{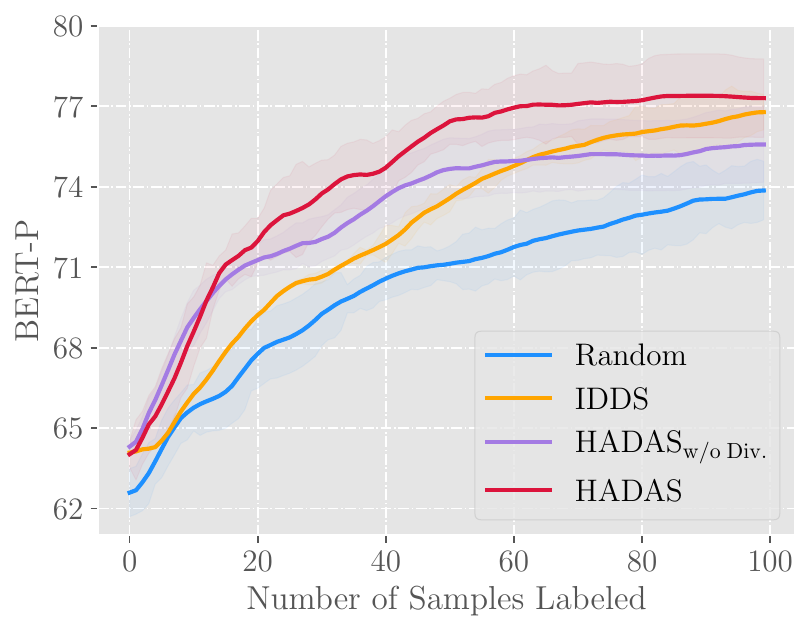}}
    \subfloat[ROUGE-L]{\includegraphics[width=0.248\textwidth]{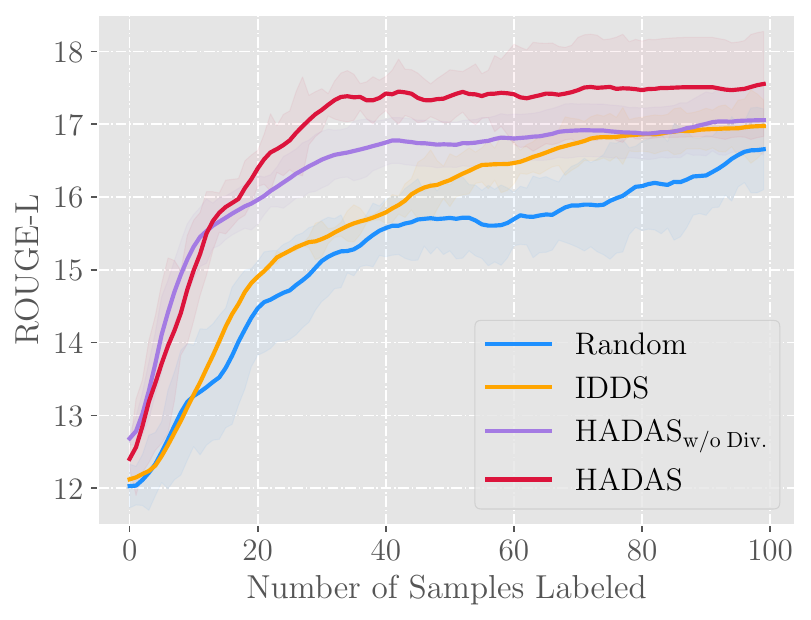}}
    \vspace{-0.5em}
    \caption{Factuality and quality curves over full hallucination annotations of Flan-T5 Base on CNN-DailyMail.}
    \label{fig:cnn-flan-t5-base} 
    \vspace{-0.5em}
\end{figure*}

\subsection{Baselines and Variants}
We compare our proposed HADAS method with the following baselines and variants.
\textbf{Random}: A canonical active learning baseline that randomly selects from samples without requiring any additional information.
\textbf{IDDS} \cite{tsvigun2022active}: A recent diversity-based sampling strategy considering semantic similarities between documents for text summarization.
\textbf{HADAS$_{\mathrm{\textbf{w/o\;Div.}}}$}: A variant of our proposed method that do not consider hallucination diversity.
\textbf{HADAS$_{\mathrm{\textbf{w/\;S.F.}}}$}: A variant of our proposed method based solely on semantic frame scores.
\textbf{HADAS$_{\mathrm{\textbf{w/\;Disc.}}}$}: A variant of our proposed method based solely on discourse scores.
\textbf{HADAS$_{\mathrm{\textbf{w/\;C.V.}}}$}: A variant of our proposed method based solely on content verifiability scores.

For the hyperparameters of HADAS and HADAS$_{\text{w/o\;Div.}}$, we set $w_1 = w_2 = w_3 = 0.33$ assuming their losses contribute equally to the hallucination generation \cite{multiloss2021}. For HADAS specifically, we did a grid search on the value of $\lambda$ across $[0.25, 0.33, 0.5, 0.67, 0.75]$ and found $\lambda = 0.33$ would yield good performance across most models and datasets.

\subsection{Active Learning Setting}\label{sec:al_setting}

As mentioned in Section \ref{sec:intro}, a notable difference between traditional NLP tasks such as NER and the hallucination mitigation we are considering is the difficulty of annotation. Annotating for hallucination is far more challenging than annotating for NER or other classification tasks.
In hallucination mitigation, there is no clear standard of what is correct or incorrect, making hallucination annotation a highly demanding task for annotators~\cite{zhao2023survey, guerreiro2023hallucinations}.
With this consideration in mind, we design a low-resource active learning setting similar to \citet{tsvigun2022active} and \citet{jukic-snajder-2023-parameter} that models the difficulty of obtaining human annotations for hallucination, thereby approximating a practical scenario.

Specifically, in each active learning iteration, only 1 sample from the unlabeled pool will be selected and annotated, which is approximately 0.05\% of the total data samples. Following the convention of previous active learning works on annotation emulation \cite{shen2018deep, ein-dor-etal-2020-active, radmard2021subsequence, shelmanov-etal-2021-active, tsvigun2022active}, we use the ground truth, i.e., gold summaries in text summarization, to emulate the human-annotated samples. 
After annotation, the model is finetuned with the annotated sample.
Note that we use standard supervised finetuning here with selected samples in each step.
Following \citet{radmard2021subsequence}, we then evaluate the model on a validation set and load the previous optimal model weights if the performance decreases after finetuning. The model is then evaluated on the test set and the performance is recorded. 
Following \citet{tsvigun2022active}, we run the active learning loop for 100 iterations with 100 annotations in total for each experiment. We use an AdamW optimizer with a learning rate of 5e-5. All experiments run on 8$\times$RTX2080Ti GPUs and are repeated 5 times. 

\section{Results}\label{sec:results}

\subsection{Performance Comparison}

\begin{table*}[!t]
\centering
\fontsize{7.21pt}{8.6pt}\selectfont

\tabcolsep=0.105cm
\renewcommand{\arraystretch}{1.3} 
\begin{tabular}{ll|cccc|cccc|cccc}
\hline
& & \multicolumn{4}{c}{\textbf{CNN-DailyMail}} & \multicolumn{4}{c}{\textbf{Multi-News}} & \multicolumn{4}{c}{\textbf{Gigaword}} \\
\cline{3-14}
\textbf{Model} & \textbf{Method} & BERT-P & UniEval & FactKB & ROUGE-L & BERT-P & UniEval & FactKB & ROUGE-L & BERT-P & UniEval & FactKB & ROUGE-L \\
\hline
\hline
\multirow{4}{*}{\begin{tabular}{@{}c@{}}Flan-T5 \\ Small\end{tabular}}
& HADAS$_{\text{w/\;S.F.}}$ & 76.96 & 72.34 & \underline{84.23} & 15.81 & 69.82 & \underline{53.98} & \underline{71.37} & 10.72 & 55.57 & 36.04 & \underline{9.03} & 21.92 \\
& HADAS$_{\text{w/\;Disc.}}$ & 76.62 & \underline{73.91} & 83.14 & 16.45 & 67.80 & 50.77 & 67.38 & 9.84 & 59.49 & \textbf{41.95} & 8.90 & \underline{23.69} \\
& HADAS$_{\text{w/\;C.V.}}$ & \underline{77.84} & 73.82 & 84.22 & \underline{16.55} & \underline{69.84} & 49.10 & 66.66 & 9.89 & \underline{60.94} & 31.88 & 8.31 & 21.40\\
& \textbf{HADAS} & \textbf{78.63} & \textbf{75.75} & \textbf{87.46} & \textbf{16.55} & \textbf{70.26} & \textbf{56.40} & \textbf{74.22} & \textbf{11.04} & \textbf{61.06} & \underline{40.53} & \textbf{10.85} & \textbf{23.89} \\
\hline 
\hline
\multirow{4}{*}{\begin{tabular}{@{}c@{}}Flan-T5 \\ Base\end{tabular}}
& HADAS$_{\text{w/\;S.F.}}$ & 72.54 & 67.08 & \textbf{80.93} & 16.64 & 68.65 & 58.75 & 62.76 & \underline{9.35} & 55.75 & 43.37 & 6.19 & 27.32 \\
& HADAS$_{\text{w/\;Disc.}}$ & 72.44 & \underline{67.64} & 75.19 & \underline{16.92} & \underline{69.28} & \underline{60.59} & \underline{65.14} & 9.18 & 58.16 & 41.70 & \underline{8.88} & \underline{29.14} \\
& HADAS$_{\text{w/\;C.V.}}$ & \underline{72.56} & 66.68 & 79.34 & 15.97 & 69.26 & 53.53 & 60.43 & 8.97 & \underline{58.27} & \underline{44.74} & 7.66 & 27.09 \\
& \textbf{HADAS} & \textbf{73.74} & \textbf{70.31} & \underline{80.73} & \textbf{17.19} & \textbf{70.82} & \textbf{61.12} & \textbf{66.39} & \textbf{9.87} & \textbf{59.36} & \textbf{47.98} & \textbf{9.18} & \textbf{29.25} \\
\hline 
\hline
\multirow{4}{*}{\begin{tabular}{@{}c@{}}BART \\ Base\end{tabular}}
& HADAS$_{\text{w/\;S.F.}}$ & 76.85 & 74.02 & \underline{93.00} & \underline{19.41} & 67.35 & 53.15 & \textbf{81.47} & \underline{13.10} & \underline{85.39} & 55.37 & \underline{61.95} & 34.43 \\
& HADAS$_{\text{w/\;Disc.}}$ & 76.68 & \underline{76.42} & 92.39 & 17.47 & 68.41 & \underline{53.63} & 70.55 & 11.33 & 82.98 & 60.27 & 53.69 & \underline{35.65} \\
& HADAS$_{\text{w/\;C.V.}}$ & \underline{76.86} & 71.61 & 86.59 & 19.38 & \underline{69.58} & 52.69 & 72.28 & 12.37 & 78.47 & \underline{61.60} & 45.19 & 34.35 \\
& \textbf{HADAS} & \textbf{78.14} & \textbf{76.65} & \textbf{93.95} & \textbf{20.12} & \textbf{71.03} & \textbf{55.94} & \underline{80.22} & \textbf{14.83} & \textbf{87.59} & \textbf{63.75} & \textbf{70.12} & \textbf{35.91} \\
\hline 
\end{tabular}
\renewcommand{\arraystretch}{1.0}
\caption{Ablation results of summarization factuality and quality metrics with 30\% of hallucination annotations across models and datasets, where the best results are highlighted in bold, and the second-best are underscored.}
\label{tab:ablation}
\vspace{-0.5em}
\end{table*}

    


We present the main evaluation results in Table \ref{tab:main}, using 30\% of the annotation budget to assess methods in a low-resource setting, accounting for the challenge of annotating hallucinations. 
Results utilizing the full annotation budget are presented in Figures \ref{fig:cnn-flan-t5-small} and \ref{fig:cnn-flan-t5-base} and discussed in \ref{sec:efficiency}.

As shown in Table \ref{tab:main}, HADAS consistently achieves the best results in hallucination evaluation metrics, spanning three different types, across all metrics and datasets. 
It also maintains high summarization qualities as measured by ROUGE-L. 
This demonstrates the effectiveness of our hallucination diversity-aware sample selection strategy. 
We also observe that, while IDDS shows a consistent advantage over the random baseline, its improvements are modest compared to those of HADAS.  
Moreover, the variant of our proposed method, HADAS$_{\text{w/o\;Div.}}$, shows clear improvements on CNN-DailyMail. However, it does not consistently outperform IDDS on the other two datasets, and it even performs worse than the random baseline in Multi-News with the BART model.
We attribute the unsatisfying performance to the greedy strategy of selecting samples that do not adequately cover the different hallucination types. 
We attribute this unsatisfying performance to the greedy strategy of selecting samples that do not adequately cover different hallucination types. 
As a result, LLMs may not comprehensively encounter cases prone to hallucination during the finetuning process. 
This underscores the importance of considering hallucination diversity in HADAS.

\subsection{Efficiency Comparison}\label{sec:efficiency}
In Figure \ref{fig:cnn-flan-t5-small} and \ref{fig:cnn-flan-t5-base}, we present the performance curves based on full hallucination annotations. Due to limited space, we only show representative curves for Flan-T5 Small and Flan-T5 Base on the CNN-DailyMail dataset and similar trends are observed in other experiments.

From Figure \ref{fig:cnn-flan-t5-small} and \ref{fig:cnn-flan-t5-base}, we observe that HADAS's performance increases rapidly in the early stages, indicating that it selects more informative hallucination samples.
Although most methods converge to comparable performance levels with more annotations, the swift improvement of HADAS underscores the efficiency of our method.
This efficiency is particularly valuable in practical applications, given the high costs and challenges associated with hallucination annotations.
Additionally, we note that while HADAS$_{\text{w/o\;Div.}}$ also shows quick initial growth, its pace slows down, and it eventually gets outperformed by IDDS with more annotations.
This phenomenon suggests that while a greedy selection may be beneficial in the short term, it might not lead to better outcomes in the long run, emphasizing the importance of considering the diversity of hallucination samples.

\subsection{Ablation Studies}
To further demonstrate the effectiveness of considering hallucination diversity, we conducted ablation experiments evaluating HADAS's performance when measuring only a single type of hallucination.
The results, presented in Table \ref{tab:ablation}, clearly show that focusing on a single hallucination type does contribute to reducing that specific type of hallucination. 
For instance, HADAS$_{\text{w/\;S.F.}}$ mostly achieves the best or second-best performances on FactKB, as it specifically targets semantic frame errors. 
Similar patterns are observed for HADAS$_{\text{w/\;Disc.}}$ and HADAS$_{\text{w/\;C.V.}}$.
However, these singular measurements alone are not sufficient for comprehensive hallucination mitigation, as some performances are even worse than the random baseline in addressing different types of hallucinations.
These ablation results further highlight HADAS's advantage in considering hallucination diversity during sample selection, consistently achieving most of the best performances across all metrics.

\section{Conclusion}
In this work, we propose the first active learning framework to mitigate hallucinations in LLMs, reducing the need for intensive human annotation efforts. 
By measuring various types of hallucinations in text summarization and developing a novel hallucination diversity-aware sample selection method, we effectively and efficiently mitigate LLM hallucinations in summarizations in a comprehensive manner.
Extensive experiments on several datasets and backbone models demonstrate the advantages of our method across various factuality metrics while maintaining high summarization quality.

\section*{Limitations}
Despite the promising results, our proposed method depends on existing hallucination detection methods to identify diverse hallucinations. 
The selection of appropriate hallucination detection metrics requires extra attention to ensure they can effectively capture various types of hallucinations.
As we discussed in Section \ref{sec:halu_measurements}, we have selected three types of hallucination detection models based on empirical results from previous works. 
However, these models may not be perfectly suited for our purposes in detecting specific hallucination types. 

Our primary contribution is the development of a generic active learning framework for hallucination mitigation, which offers the flexibility to easily integrate additional hallucination detection methods. 
We plan to conduct more comprehensive experiments using more fine-grained and interpretable hallucination detection methods in future work.

Additionally, in our experiments, we followed the practices of prior active learning studies by using ground-truth data to emulate human annotations, specifically gold summaries in our context. However, recent works suggest that these gold summaries might also contain hallucinated content. Although we have intentionally chosen datasets with more reliable gold summaries, conducting experiments with actual human annotations would be highly beneficial to further evaluate the effectiveness of our active learning framework.

\section*{Ethical Consideration}
Active learning inherently involves biased sampling, which can potentially result in datasets with biased annotations. Consequently, this approach can be intentionally employed to amplify existing biases within datasets. Our research enhances the effectiveness of hallucination mitigation, thereby also increasing its capacity to introduce hallucinations more efficiently. Therefore, extra cautions are needed for any practical application of our method.

\bibliography{anthology,custom}

\appendix

\balance
\begin{table*}[!t]
\centering
\fontsize{7.21pt}{8.6pt}\selectfont
\tabcolsep=0.10cm
\renewcommand{\arraystretch}{1.3} 
\begin{tabular}{ll|cccc|cccc|cccc}
\hline
& & \multicolumn{4}{c}{\textbf{CNN-DailyMail}} & \multicolumn{4}{c}{\textbf{Multi-News}} & \multicolumn{4}{c}{\textbf{Gigaword}} \\
\cline{3-14}
\textbf{Model} & \textbf{Method} & BERT-P & UniEval & FactKB & ROUGE-L & BERT-P & UniEval & FactKB & ROUGE-L & BERT-P & UniEval & FactKB & ROUGE-L \\
\hline
\hline
\multirow{4}{*}{\begin{tabular}{@{}c@{}}LLaMa-2 \\ 7B\end{tabular}}
& Random & 57.72 & 52.42 & 75.44 & 14.28 & 55.69 & 62.02 & 60.74 & 10.33 & 60.78 & 58.28 & 21.38 & 21.13 \\
& IDDS & \underline{58.61} & 58.45 & \underline{79.80} & \underline{16.15} & 53.73 & 61.04 & 61.08 & \textbf{11.35} & 63.20 & \underline{60.61} & 21.12 & \underline{23.77}  \\
& HADAS$_{\text{w/o\;Div.}}$ & 58.19 & \underline{58.90} & 78.84 & 15.58 & \underline{58.67} & \underline{63.82} & \underline{61.73} & 10.57 & \underline{63.56} & 60.12 & \underline{24.02} & 23.49  \\
& \textbf{HADAS} & \textbf{60.34} & \textbf{60.61} & \textbf{83.52} & \textbf{16.75} & \textbf{58.79} & \textbf{64.28} & \textbf{64.89} & \underline{10.93} & \textbf{64.77} & \textbf{62.35} & \textbf{24.26} & \textbf{23.81}  \\
\hline 
\end{tabular}
\renewcommand{\arraystretch}{1.0}
\caption{Preliminary results of summarization factuality and quality metrics with 30\% of hallucination annotations of LLaMa-2 7B on all datasets, where the best results are highlighted in bold, and the second-best are underscored.}
\label{tab:main_llama}
\vspace{-0.5em}
\end{table*}

\section{LLaMa-2 Experiments}\label{app:llama}

We also conduct some experiments on the LLaMa-2 7B to show the effectiveness of our method on larger models in mitigating hallucinations.
Similarly in Section \ref{sec:models}, we first finetune the model on the XSum dataset to ensure summarization quality. 
Specifically, we use LoRA \cite{hu2022lora} finetuning with a randomly selected subset of 5000 samples with the prepended instruction prompt "\texttt{Summarize the following article:}". 
The LoRA rank is set to 8 and alpha is set to 16. An AdamW optimizer is used with a learning rate of 5e-4.
We then use the LLaMa-2 7B model finetuned on the XSum dataset as the initial point for active learning. 
The same instruction prompt is used for experiments on all three datasets.
The rest of the active learning settings, dataset construction, and hyperparameter selection in the experiments remain the same as detailed in Section \ref{sec:experiment}, except that we also apply LoRA finetuning with the above configuration during the active learning process. 
The experiments run on 2$\times$A40 GPUs and are repeated 3 times. The results are shown in Table \ref{tab:main_llama}.

From Table \ref{tab:main_llama}, we observe that HADAS consistently outperforms baselines on most evaluation metrics. 
The results again validate advantages of our method as similarly observed in Section \ref{sec:results}. 
Note that in these experiments we do not optimize the training configurations and parameters, which may influence the LoRA finetuning performances of LLaMa-2 models as suggested by \citet{li2024loftq} and lead to sub-optimal results. 
We leave more comprehensive evaluations of our hallucination mitigation method on LLMs of larger parameter sizes as future work.

\end{document}